%% file: eccv2022submissionCR.tex
\documentclass[runningheads]{llncs}

\usepackage{graphicx}
\usepackage{orcidlink}
\usepackage{tikz}
\usepackage{amsmath,amssymb} 
\usepackage{color}
\usepackage{booktabs}
\usepackage{multirow}
\usepackage{appendix}
\usepackage{pdfpages}
\usepackage{caption}
\usepackage{subcaption}
\usepackage{marvosym}
\usepackage{enumitem}                                   
\usepackage{chngcntr}                                   
\usepackage{hyperref}
\hypersetup{
     colorlinks=true,
     linkcolor=black,
     filecolor=black,
     citecolor = black,      
     urlcolor=black,
     }

\usepackage[accsupp]{axessibility}

\usepackage{bm}

\begin{document}
\pagestyle{headings}
\mainmatter

\title{Label-Guided Auxiliary Training Improves\\3D Object Detector} 
\titlerunning{Label-Guided Auxiliary Training Improves 3D Object Detector}

\author{Yaomin Huang \inst{1}\textsuperscript{,*}\orcidlink{0000-0002-8195-4978} \and
Xinmei Liu \inst{1}\textsuperscript{,*}\orcidlink{0000-0001-7649-8085} \and
Yichen Zhu \inst{2}\orcidlink{0000-0001-5126-838X}\and
Zhiyuan Xu \inst{2}\orcidlink{0000-0003-2879-3244}\and \\
Chaomin Shen \inst{1}\textsuperscript{,\Letter}\orcidlink{0000-0001-9389-6472} \and
Zhengping Che \inst{2}\orcidlink{0000-0001-6818-1125} \and
Guixu Zhang \inst{1}\and
Yaxin Peng \inst{3}\orcidlink{0000-0002-2983-555X}\and \\
Feifei Feng \inst{2}\and
Jian Tang \inst{2}\textsuperscript{,\Letter}\orcidlink{0000-0003-4418-0114}
}

\authorrunning{Huang, Liu, et al.}
\institute{School of Computer Science, East China Normal University \and
AI Innovation Center, Midea Group \and
Department of Mathematics, School of Science, Shanghai University\\
\email{\{51205901049,51205901078\}@stu.ecnu.edu.cn} \\
\email{\{cmshen,gxzhang\}@cs.ecnu.edu.cn} \\
\email{\{zhuyc25,xuzy70,chezp,feifei.feng,tangjian22\}@midea.com}\\
\email{yaxin.peng@shu.edu.cn}
\footnote[0]{\textsuperscript{*}~Equal contributions; work done during internships at Midea Group.
\\ \textsuperscript{\Letter}~Corresponding authors.}
}

\maketitle

\begin{abstract}
Detecting 3D objects from point clouds is a practical yet challenging task that has attracted increasing attention recently. In this paper, we propose a Label-Guided auxiliary training method for 3D object detection (LG3D), which serves as an auxiliary network to enhance the feature learning of existing 3D object detectors. Specifically, we propose two novel modules: a Label-Annotation-Inducer that maps annotations and point clouds in bounding boxes to task-specific representations and a Label-Knowledge-Mapper that assists the original features to obtain detection-critical representations. The proposed auxiliary network is discarded in inference and thus has no extra computational cost at test time. We conduct extensive experiments on both indoor and outdoor datasets to verify the effectiveness of our approach. For example, our proposed LG3D improves VoteNet by 2.5\% and 3.1\% mAP on the SUN RGB-D and ScanNetV2 datasets, respectively.
The code is available at \url{https://github.com/FabienCode/LG3D}.
\end{abstract}

\section{Introduction}
3D object detection is one of the fundamental tasks toward precisely and adaptively understanding the real 3D world.
Specifically, 3D object detection processes point clouds (as shown in Fig.~\ref{fig:pc}) to identify the types of objects and localize their bounding boxes (as shown in Fig.~\ref{fig:bb}).
While challenging and computationally expensive, 3D object detection has attracted wide attention with an increasing amount of excellent works~\cite{chen2020hierarchical,chen2020object,cheng2021back,liu2021group,misra2021end,qi2020imvotenet,qi2019deep,qi2018frustum,DBLP:conf/iccvw/WangZJCTSP21,xu2021spg,yoo20203d,zhang2020h3dnet,zheng2021se}. Existing 3D object detection methods mostly focus on improving the feature extraction in point clouds and making better predictions on objects' locations,
such as fusing 2D image and 3D data information~\cite{qi2020imvotenet}, leveraging a shape attention graph convolution operator (SA-GConv)~\cite{chen2020hierarchical} to capture local shape features and relative geometric positions between points, and introducing a strong backbone for better feature learning ability~\cite{liu2021group}.
Nevertheless, one of the most critical issues is that inference speed is typically sacrificed in order to maintain a high performance of 3D object detectors.

Balancing the inference speed and detection performance is challenging due to the nature of point clouds, i.e., the number of points in practical scenarios is huge, which slows down the forward pass. One can bypass such obstacles by applying aggressive sampling strategies, but it severely hurts the quality of detectors. Instead of modifying the architecture of the existing 3D object detectors, in this paper we resolve this issue by introducing a model-agnostic auxiliary training approach, which dramatically improves the detection performance and brings no extra computational cost at test time. Our proposed method is motivated by the assumption that the input labels (i.e., points within bounding boxes) contain rich semantic information if one could find a proper way to extract its latent features. These features can be considered an auxiliary information source, provide supervision to 3D object detectors during training, and, more importantly, can be removed after the training stage. As such, 3D object detectors can be optimized more effectively without hampering the inference speed.

\input{Fig/BPC}

The previous approach adopts learnable modules to extract features from labels in the 2D tasks. For instance, Mostajabi et al.~\cite{mostajabi2018regularizing} used an auto-encoder on the semantic masks to help the image segmentation model learn better pixel-level features. Similarly, LabelEnc~\cite{hao2020labelenc} and LGD~\cite{zhang2022lgd} formulate the bounding box along with its class identity as an extra source of information to supervise the student model. However, despite these previous attempts at learning label information, applying it to 3D detection is non-trivial due to the fundamental difference in input structure between 2D and 3D tasks, i.e., the image in the 2D task versus point clouds in 3D detection. Moreover, besides the categorical information, the point clouds inside the bounding box, i.e., the label point clouds shown in Fig.~\ref{fig:lpc}, contain rich semantic and position information of each target object in the scene, which have been overlooked in the prior work.

Motivated by the above analysis, in this paper, we propose a Label-Guided auxiliary training approach for 3D object detection (LG3D), which serves as an auxiliary network to enhance the feature learning ability of vanilla 3D object detectors. To better utilize the 3D label information, we introduce two novel modules in our method. First, the Label-Annotation-Inducer (LAI) module parameterizes the bounding box label and then maps them to task-specific representations. It aims to fuse the point clouds of particular objects into the sparse, original point clouds input such that the detectors can realize the object's localization, along with other critical but unexplored high-dimensional features, learned particularly by a tiny label encoder. The Label-Knowledge-Mapper (LKM) module is followed up to obtain optimal representations. Despite the simple design of our proposed modules, it tremendously improves the performance of 3D object detectors. It's also worth noting that our proposed LG3D is only used in the training stage and is completely cost-free during the inference.

We summarize our contributions as follows:
\begin{itemize} [itemsep=0pt,topsep=2pt]
\item We propose LG3D, a new way to utilize 3D labels by using the label point clouds (i.e., point clouds inside bounding boxes) as an auxiliary network to assist the feature representation learning of the vanilla network.
\item Two novel modules, LAI and LKM, are used to fuse label point clouds, annotations, and original point clouds to a single feature embedding, which can effectively compensate for the missing information of target objects caused by data sampling.
\item The proposed LG3D can be simply inserted into existing 3D object detectors and removed after training. LG3D improves the state-of-the-art 3D object detectors by a large margin on both indoor and outdoor datasets.
\end{itemize}

\section{Related works}
\label{sec:related}

\subsection{3D Object Detection}
We briefly introduce the 3D object detection approaches in this section, and refer reader to Qian et al.~\cite{qian20223d} for more detailed description. ImVoteNet \cite{qi2020imvotenet} proposes to use 2D image RGB, geometric coordinates, semantics, and pixel texture information to assist 3D point clouds object detection. PointPainting \cite{vora2020pointpainting} proposes to use 2D semantic segmentation information to fuse the transformation matrix of LiDAR information and image information to the point.
Cross-modal information fusion is proposed in the PointAugmenting \cite{wang2021pointaugmenting} method, point features of corresponding points in 2D images are extracted by mapping between 3D and 2D.
BRNet \cite{cheng2021back} proposes to solve the problem that VoteNet \cite{qi2019deep} cannot effectively represent the object structure information, adding a back-tracing module for resampling the more informative seed points. HGNet \cite{chen2020hierarchical} describes the shape of an object by simulating the relative geometric position of the point. H3DNet \cite{zhang2020h3dnet} votes for center points on three dimensions of the bounding box, bounding box surface, and bounding box edges to add more detailed constraints to bounding box predictions. The backtracking module \cite{cheng2021back} is added based on VoteNet \cite{qi2019deep} to resample the seed points with richer information.
3DSSD \cite{yang20203dssd} achieves a good balance between accuracy and efficiency by using the fusion sampling strategy in the downsampling process. In GroupFree3D \cite{liu2021group}, the transformer adaptively determines the relationship between points and obtains an object proposal by point aggregation. DETR3D \cite{wang2022detr3d} uses DETR for 3D object detection, extracting 2D features from multiple camera images, then indexing these 2D features using a set of sparse 3D target queries, using a camera transform matrix to establish connections between 3D positions and multi-view images, and finally connecting 2D feature extraction and 3D box prediction by alternating between 2D and 3D calculations.
3DETR-m \cite{misra2021end} improves detection performance by applying mask to self-attention in transformer. However, the calculation cost of object detection increases with the increment in the use of transformers.

Despite the evolutionary development of 3D object detectors, current approaches still require overwhelming computational costs at test time to maintain satisfactory performance. Thus, we provide a novel perspective to harness the semantic information in the label to assist the training of a 3D object detector, which is the first work demonstrating the powerful yet unexplored information in the point clouds that, if handled properly, can
significantly boost the existing 3D detector. Our approach is also detector agnostic and robust to different kinds of datasets.

\subsection{Auxiliary Task and Knowledge Distillation}
\subsubsection{Auxiliary Task}
Auxiliary task~\cite{zhang2020auxiliary} is a well-studied topic that aims to assist the model with a lightweight module during training or testing. For example, in SA-SSD~\cite{he2020structure}, the original point cloud features are complemented with down-sampled features with an auxiliary task. While auxiliary training in 3D detection has not raised attention, it has developed fast in 2D object detection. For instance, LabelEnc~\cite{hao2020labelenc} proposes directly introducing auxiliary intermediate supervision to the trunk to provide feasible supervision in the training stage. It is further modified~\cite{zhang2022lgd} into a teacher-free approach that incorporates bounding box and class information to the student network.

\subsubsection{Knowledge Distillation}
Knowledge distillation (KD) is another highly closed topic in our approach. It was initially proposed to leverage a large teacher network that transfers its representative knowledge to a compact student network. Its success has spread over numerous domain in computer vision, i.e., image classification~\cite{DBLP:journals/corr/HintonVD15,zhu2021student}, object detection~\cite{zhang2020improve,kang2021instance}, semantic segmentation~\cite{liu2019structured}, and image-to-image translation~\cite{zhang2022region,zhang2022wavelet}.

For 3D object detection, SE-SSD \cite{zheng2021se} uses the idea of knowledge distillation to optimize student networks through a combination of hard and soft targets. Wang et al.~\cite{wang2020pad} uses KD to compensate for the gap between the model of training high-quality input and the model of testing low-quality input in reasoning. Chong et al.~\cite{chong2022monodistill} leverage point clouds to assist monocular 3D object detection with depth information. More recently, PointDistiller~\cite{zhang2022pointdistiller} leverages the dynamic graph convolution to transfer the local geometric structure of point clouds.

This work combines two advantages in the auxiliary task and knowledge distillation. Namely, our approach does not require a heavy, cumbersome teacher model to perform distillation. At the same time, we still enjoy the improvement in performance without extra computational cost at test time, which is normally unavoidable in training with the auxiliary task.

\section{Method}

\begin{figure}[t]
\centering
\includegraphics[width=.95\textwidth]{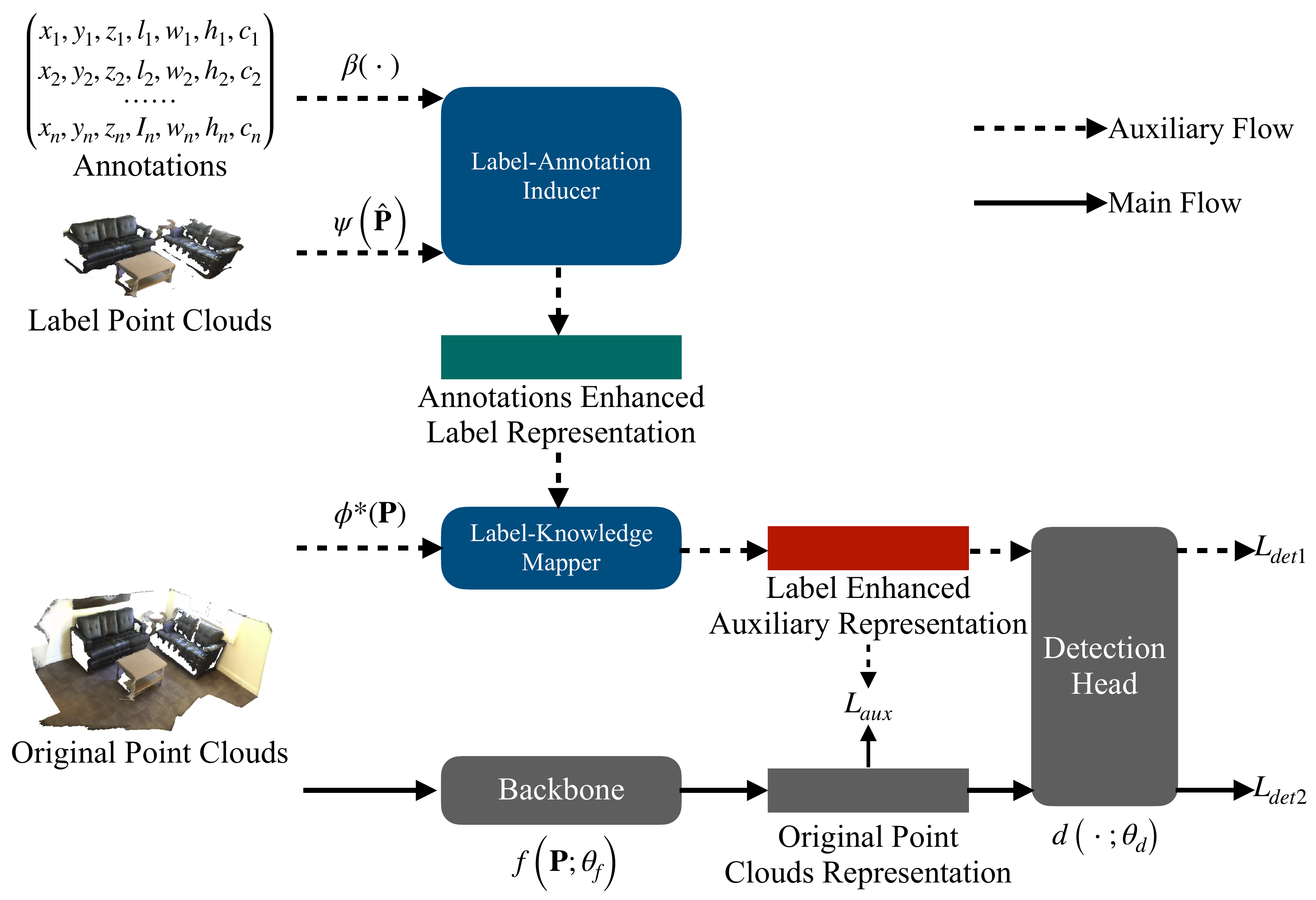}
\caption{Our LG3D framework. It includes an LKM module and an LAI module. The whole framework can be simply inserted into a 3D object detection network, and the LKM module shares the detection head with the backbone network. LG3D is removed directly in the inference phase, so it does not increase the computational cost. As shown in the figure, data flow in the training stage contains dotted and solid arrows, while data flow in the inference stage only contains solid arrows.}\label{fig:Framework}
\end{figure}

In this section, we present our method in detail. Fig.~\ref{fig:Framework} gives an overview of our method. In Sec~\ref{sec:3.1}, we introduce LKM supplements the original point clouds representation with label point clouds to obtain the label enhanced auxiliary representation. In Sec \ref{sec:3.2},  LAI encodes the annotations and maps it to a latent semantic space for the annotation enhanced label representation that aim to get a better label enhanced auxiliary representation. In Sec \ref{sec:3.3}, a separable auxiliary task uses the label enhanced auxiliary representation to supervise the representation outputs from the backbone with original point clouds.

\subsection{Label-Knowledge Mapper}\label{sec:3.1}

\begin{figure}[t]
\centering
\includegraphics[width=.98\textwidth]{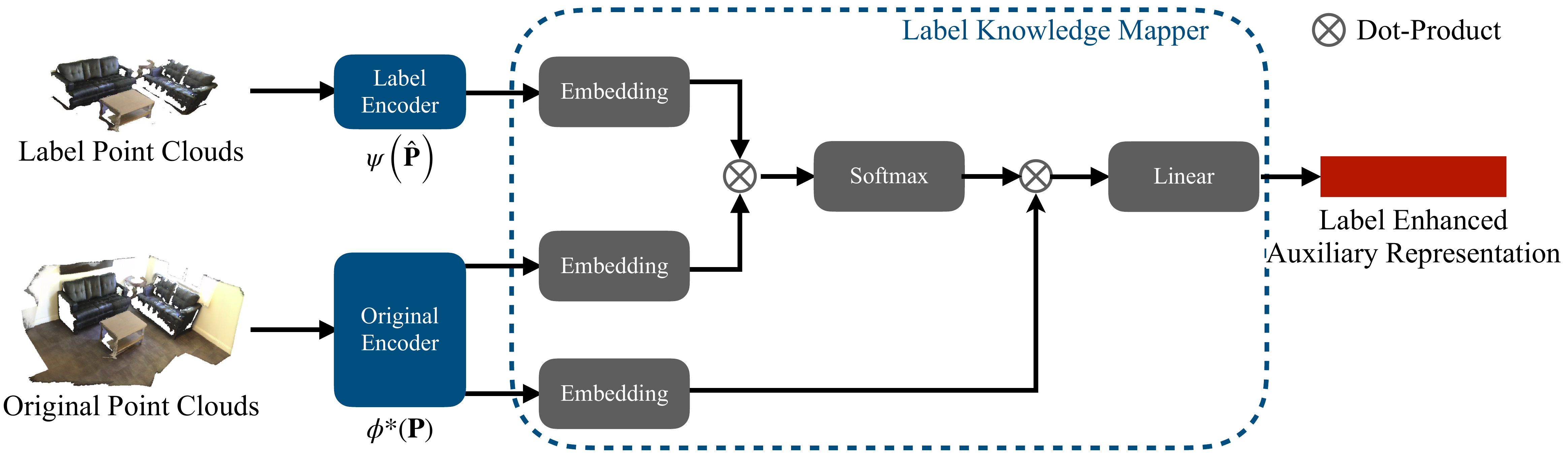}
\caption{The dashed box indicates the entire LKM module.}\label{fig:LKM}
\end{figure}

The features of the original point clouds are usually extracted by the sampling method. This way, more or fewer point clouds of the object will be lost, affecting the feature extraction. Thus we design the LKM module to induce instance features from label point clouds, then fuse it with the original point clouds representation. This module can well supplement the key information lost during the point clouds samplings, especially for the information loss of small objects. 

Original point clouds and label point clouds can be represented as disordered point set $\bm{P}=\left\{{\bm{p}}_{i}\right\}_{i=1}^{n}$ with $\bm{p}_{i} \in \mathbb{R}^{d}$ and $\bm{\hat{P}}=\left\{\hat{\bm{p}_{j}}\right\}_{j=1}^{m}$ with $\hat{\bm{p}_{j}} \in \mathbb{R}^{d}$ respectively, 
where $n$ and $m$ is the number of original point clouds and label point clouds, respectively. 
$d$ represents the $(x,y,z)$ coordinate plus extra feature channels such as color, normal, etc.
As shown in Fig. \ref{fig:LKM}, in the training stage, we obtain the label point clouds by annotation information in the dataset and feed it into the label encoder.
Given a training set $(\bm{P}, \bm{\hat{P}})$ and a well-trained original encoder function $\phi^{*}(\bm{P})$, instead of fine-tuning the original feature representation
to the task-specific label space, we fix $\phi^{*}(\cdot)$ and learn a separate label encoder $\psi(\bm{\hat{P}})$ to extract the feature from label point clouds.
Then we use a label fusion function
$\mathcal{H}((\bm{P},\bm{\hat{P}}),\theta_{\mathcal{H}})$
to fuse the original point clouds and the label point clouds.
We find the optimal representation and function by
\begin{align}\label{equ1}
\begin{aligned}
\theta_{f}^{*}, \theta_{d}^{*}=\underset{\theta_{f}, \theta_{d}}{\arg \min \ }
\mathbb{L}_{\text{det}}^{1}
(d(\mathcal{H}((\bm{P}, \hat{\bm{P}}), \theta_{\mathcal{H}}) ; \theta_{d}), y)
& +\mathbb{L}_{\text{det}}^{2}(d\left(f\left(\bm{P} ; \theta_{f}\right) ; \theta_{d}\right), y) \\
&+\lambda \mathbb{L}_{\text{aux}},
\end{aligned}
\end{align}
where $y \in \mathbb{R}^{N\times V}$ is the ground-truth label, $N$ is the number of objects, and $V$ is the label length of each objects.
$f\left(\bm{P} ; \theta_{f}\right)$ is the function realized by the backbone. $\mathcal{H}((\bm{P},\bm{\hat{P}}),\theta_{\mathcal{H}}) \in \mathbb{R}^{n'\times C}$ represents the output of the LKM module, where $n'$ is the number of sampled points and $C$ is the number of feature channels.
$\mathbb{L}_{\text{aux}}$ represents the auxiliary loss attached to the outputs of the backbone, which is independent of the detection head $d\left(\cdot, \theta_{d}\right)$ thus it is not affected by the latter’s convergence progress.
$\lambda$ is the balanced coefficient.

The design of $\mathbb{L}_{\text{aux}}$ is one of the most important factor of our approach and we will explain its design in Sec~\ref{sec:3.3}.
$\mathbb{L}_{\text{aux}}$ aims to minimize the distance between the original point clouds feature representation and an ideal representation, which in our method is the label enhanced auxiliary representation. In order to make sure that the original point clouds features can be well combined with the features of the label point clouds, we adopt the attention mechanism\cite{vaswani2017attention}.
Let the matrix representations of the key ($\bm{K}_l$) and value ($\bm{V}_l$) be
$\phi^{*}(\bm{P} ; \theta_{p}) \in \mathbb{R}^{M\times C}$ from the original point clouds representation,
query be $\bm{Q}_{l} = \psi(\bm{\hat{P}} ; \theta_{\hat{P}}) \in \mathbb{R}^{M\times C}$
, with the label point clouds representation. Here $M$ and $C$ denotes length and dimensions of query, key and value, respectively.
The query, key and value are transformed by linear layers $f_{\mathcal{Q}_l}$, $f_{\mathcal{K}_l}$, $f_{\mathcal{V}_l}$
before conducting attention.
To induce the feature from label point clouds, we apply the cross-attention mechanism~\cite{vaswani2017attention} to fetch original point clouds representation from label point clouds representation. So the ideal representation output of LKM is:

\begin{equation}
\mathcal{H}\left((\bm{P},\bm{\hat{P}}),\theta_{\mathcal{H}}\right)=\operatorname{Softmax}\left(\frac{{f_{\mathcal{Q}_{l}}(\bm{Q}_{l})f_{\mathcal{K}_{l}}(\bm{K}_l)}^{\top}}{\sqrt{D_{k}}}\right) f_{\mathcal{V}_{l}}(\bm{V}_{l}),
\label{attention_epu}
\end{equation}
where $D_{k}$  denotes the dimensions of the key, and $\operatorname{Softmax}(\cdot)$ is applied row-wise.

\subsection{Label-Annotation-Inducer}\label{sec:3.2}

\begin{figure}[t]
\centering
\includegraphics[width=.98\textwidth]{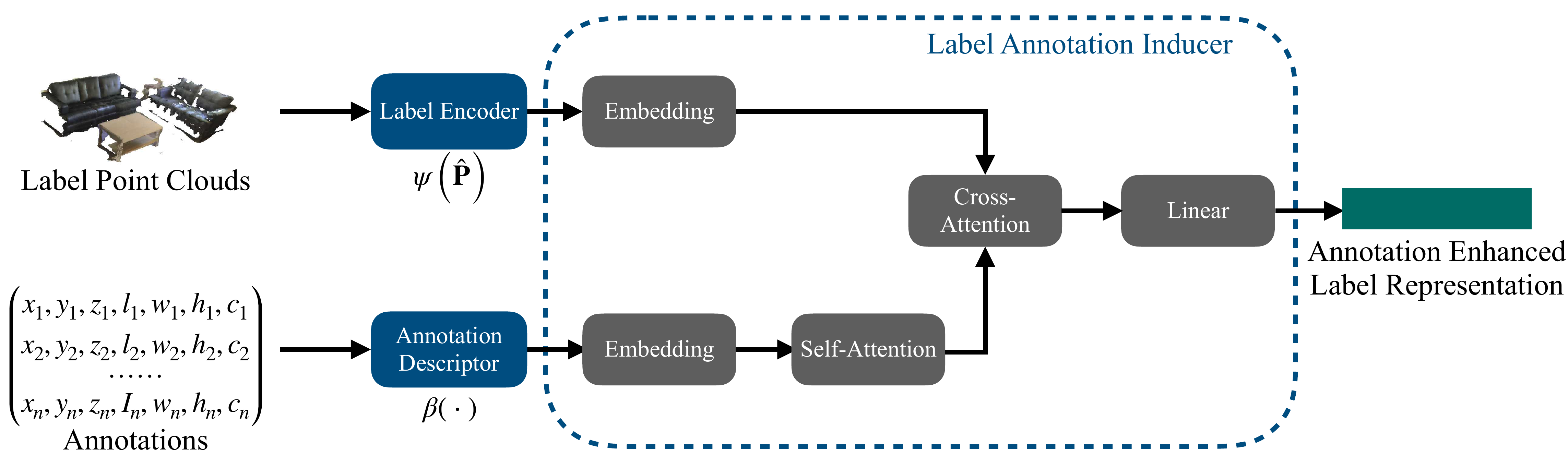}
\caption{In the training phase, label point clouds and annotations are fed into the LAI module.}
\label{fig:LAI}
\end{figure}

Using the proposed Label-Knowledge Mapper described in Sec.~\ref{sec:3.1}, we obtain the enhanced point clouds representation $\mathcal{H}$.
However,
the rich information in the annotations has not been fully utilized. To use the label annotations information as an important form of ground truth, we propose the LAI module, as shown in Fig.~\ref{fig:LAI}, to complement the features of the label point clouds. Specifically,
we extract the label annotations information to obtain an ideal representation $\mathcal{G}$.

\subsubsection{Label Embedding}
In a 3D object detection task, the label information of an object usually contains the center, the size and category, and it may have the head angle. Given an object label, we represent each labeled bounding box of the target object in the point clouds as $\bm{\alpha}_{i} = ( x_i, y_i, z_i, l_i, w_i, h_i, c_i )$, where $i$ represents the $i$-th bounding box, $(x_i, y_i, z_i)$ represents the center point of the $i$-th bounding box, $(l_i, w_i, h_i)$ represents the length, width and height of the $i$-th bounding box, and $c_i$ represents the object category corresponding to the $i$-th bounding box. The initial label representation is
\begin{equation}
\mathcal{A}=\left\{\bm{\alpha}_{1} , \cdots, \bm{\alpha}_{i}, \cdots , \bm{\alpha}_N\right \},
\quad\bm{\alpha}_i \in \mathbb{R}^{C_L},
\end{equation}
where $i$ indicates the object index, $C_{L}$ is the array length of the bounding box parameter information
and $N$ is the object number.

\subsubsection{Annotation Augmentation}
We perform some information dropping for these determined annotations. When describing bounding boxes, we interpret them with rough scale indices. The centers of the approximate fields $( x_i^{\prime}, y_i^{\prime}, z_i^{\prime})$ are obtained by random dithering as
\begin{equation}
\begin{aligned}
&x_{i}^{\prime}=x_{i}+\eta_{x} l_{i}, \\
&y_{i}^{\prime}=y_{i}+\eta_{y} w_{i}, \\
&z_{i}^{\prime}=z_{i}+\eta_{z} h_{i},
\end{aligned}
\end{equation}
where $\eta_{x}$, $\eta_{y}$, $\eta_{z}$ are sampled from a uniform distribution $\eta \sim U[-B_\eta, B_\eta]$, where $B_\eta$ is a scale factor whose value is set to be 0.1.
Furthermore, we generate fake instances for the recognition task based on the dataset distribution. Note that we need to identify the objectiveness of each instance, that is, to determine the authenticity of a given label. We use the binary cross-entropy function to determine whether the given label is a real label in the point clouds scene or a pseudo label that we add manually and increase the robustness of learning knowledge from the real label by determining the virtual label:
\begin{equation}
\mathbb{L}_{\text{idf}}=-\frac{1}{N} \sum_{i=1}^{N} \delta_{\text{obj}}\left(\bm{\alpha}_{i}\right) \log \left(\mathcal{P}_{\text{obj}}\left(\bm{e}_{i}\right)\right)+\left(1-\delta_{\text{obj}}\left(\bm{\alpha}_{i}\right)\right) \log \left(1-\mathcal{P}_{\text{obj}}\left(\bm{e}_{i}\right)\right),
\end{equation}
where $\mathcal{P}_{\text{obj}}(\cdot)$ is a prediction function with a full connection layer and sigmoid function, and $\delta_{\text{obj}}(\bm{\alpha}_{i})$ encodes the binary classification labels, denoting whether the instance is randomly generated ($\delta_{\text{obj}}\left(\bm{\alpha}_{i}\right) = 0$) or manually annotated ($\delta_{\text{obj}}\left(\bm{\alpha}_{i}\right) = 1$).

We then introduce the annotations descriptor $\beta(\cdot)$, which could induce the new label representation $\mathcal{A}$ to the task-specific latent feature space. We adopt a relatively simple multi-layer perceptron \cite{hastie2009elements} as the label annotation encoding module, so the optimization seems not difficult. The new label annotation representation is:
\begin{equation}
\beta_{\mathcal{A}}=\left\{\bm{e}_{1}, \cdots, \bm{e}_{i}, \cdots , \bm{e}_{N}\right\},
\quad\bm{e}_i \in \mathbb{R}^C,
\end{equation}
where $C$ is the intermediate feature dimension, and $\bm{e}_i = \beta(\bm{\alpha}_{i})$ is the encoded label annotation information.

\subsubsection{Label Information Interactions}
Since the annotation representation $\beta_{\mathcal{A}}$ is relatively independent, we first model it globally by a self-attention to obtain the global annotation representation $\bm{Q}_{\alpha} \in \mathbb{R}^{N\times C}$. We then use this as a query condition to apply the cross attention mechanism to the label point clouds' features, so that the label point clouds can be combined with the annotation information to produce better annotation enhanced label representation.

Specifically, given the matrix representations of query $\bm{Q}_{\alpha} \in \mathbb{R}^{M\times C}$, key $\bm{K}_{\alpha} \in \mathbb{R}^{M\times C}$ and value $\bm{V}_{\alpha} \in \mathbb{R}^{M\times C}$ are from the label point clouds representation. Before conducting cross attention, the query, key, and value are transformed by linear layers $f_{\mathcal{Q}_{\alpha}}$, $f_{\mathcal{K}_{\alpha}}$, $f_{\mathcal{V}_{\alpha}}$,

\begin{equation}\label{eq8}
\mathcal{F}_{A}\left(\bm{Q}_{\alpha}, \bm{K}_{\alpha}, \bm{V}_{a}\right)=\operatorname{Softmax}\left(\frac{{f_{\mathcal{Q}_{a}}(\bm{Q}_{\alpha})f_{\mathcal{K}_{\alpha}}(\bm{K}_{\alpha})}^{\top}}{\sqrt{D_{k}}}\right) f_{\mathcal{V}_{\alpha}}(\bm{V}_{\alpha}).
\end{equation}

With the LAI making label point clouds representation perceive label annotations information, the new label fusion function is:
\begin{equation}\label{equgfun}
\mathcal{G}\left(\bm{P}, (\hat{\bm{P}}, \mathcal{A});\theta_\mathcal{G}\right)
=\mathcal{G}\left(\phi^{*}\left(\bm{P} ; \theta_{P}\right), \left(\psi(\hat{\bm{P}} ; \theta_{\hat{P}}),  \beta(\mathcal{A};\theta_{\mathcal{A}})\right)\right).
\end{equation}

\subsection{Separable Auxiliary Tasks}\label{sec:3.3}
In Sec. \ref{sec:3.1}, we propose to use label point clouds to supplement the original point clouds representation. In Sec. \ref{sec:3.2}, the label point clouds representation is further enriched by label annotations information. After these modules, we obtain an ideal representation $\mathcal{G}$ which can supervise the representation outputs from the backbone with original point clouds. We propose a separable auxiliary network using the above modules. It can be simply insert into various 3D object detection networks to improve the detection accuracy during the training.

It is clear that Eq.~\eqref{equ1} directly corresponds to a multi-task training paradigm with three loss terms: the first one is label information encoder loss ($\mathbb{L}_{\text{det}}^{1}$) for the label information embedding; the second term is the common detection loss ($\mathbb{L}_{\text{det}}^{2}$), which enforces $d\left(\cdot ; \theta_{d}^{\prime}\right)$ to be a valid detection head; the third loss ($\mathbb{L}_{\text{aux}}$) minimizes the gap between the two latent spaces (namely the outputs of the backbone $f(\cdot ; \theta_{f}^{\prime})$ and the label enhanced auxiliary representation $\mathcal{G}((\bm{P}, (\bm{\hat{P}}, \mathcal{A})),\theta_\mathcal{G}$).

By sharing the detection head for supervision, we ensure the instructive representation quality and consistency with the original point clouds representation. The overall detection loss is:
\begin{equation}
\begin{aligned}
\mathbb{L}_{\text{det}}
&=\mathbb{L}_{\text{det}}^{1}+\mathbb{L}_{\text{det}}^{2} + \mathbb{L}_{idf}.
\end{aligned}
\end{equation}

In addition to the common detection loss $\mathbb{L}_{\text {\text{det}}}$, we introduce an auxiliary supervision loss $\mathbb{L}_{\text{aux}}$ that uses outputs from LKM directly to supervise the detection backbone, as flow:
\begin{equation}
\mathbb{L}_{\text{aux}} =  \min _{\theta_{f}}{\left\|f\left(\bm{P},\theta_{f}\right)-\mathcal{G}\left((\bm{P}, (\bm{\hat{P}}, \mathcal{A})), \theta_\mathcal{G} \right)\right\|}_2,
\end{equation}
where ${\|\cdot\|}_2$ is L2-distance to minimize the difference between original point clouds representation and label enhanced auxiliary representation.
It is worth noticing that the gradients of $\mathbb{L}_{\text{aux}}$ only update the backbone module. Above all, the overall loss with a coefﬁcient $\lambda$ can be summarised as follows:
\begin{equation}
\mathbb{L}_{\text{total}}=\mathbb{L}_{\text{det}}+\lambda \mathbb{L}_{\text {aux}}.
\end{equation}

In summary, we use label-guided auxiliary training to motivate the underlying network to learn better feature representations. As Fig. \ref{fig:Framework} shows, during the testing phase, all dotted arrow flow lines are removed, so no additional computational overhead is incurred.

\section{Experiments}

\subsection{Experiment Settings}
\subsubsection{Dataset}
To illustrate the generalization of our method, we have conducted experiments on indoor and outdoor datasets.
For indoor datasets, the SUN RGB-D dataset consists of 10,355 single-view indoor RGB-D images annotated with over 64,000 3D bounding boxes and semantic labels for 37 categories.
The ScanNetV2 dataset is a 3D mesh dataset with about 1,500 3D reconstructed indoor scenes with 40 semantic classes.
We follow the commonly-used settings, selecting 10 classes of SUN RGB-D and 18 classes of ScanNetV2.
For outdoor datasets, we choose the KITTI dataset for evaluation. The KITTI dataset contains 7481 training samples and 7518 test samples with three categories: Car, Pedestrian and Cyclist.

\subsubsection{Data Preparation}
In the training stage, our network has two different inputs. On the one hand, we feed the full point clouds into the main branch to extract feature representation. On the other hand, we feed the point clouds inside in the bounding box and label information into the auxiliary network. The point clouds are randomly sub-sampled from the raw data of each dataset, i.e., 20,000 points from point clouds in the SUN RGB-D dataset and 40,000 point clouds from a 3D mesh in the ScanNetV2 dataset. Additionally, we perform data augmentation by randomly flipping, rotating, and scaling the point clouds.

\subsubsection{Training and Evaluation}
We implement our LG3D using MMdetection3D \cite{mmdet3d2020} framework. For different networks with different datasets, we followed the basic settings in MMdetection3D without additional parameter tuning. The evaluation for indoor datasets follows the same protocol as \cite{qi2019deep} using mean average precision mAP@0.25 and mAP@0.50.
We only evaluate our model on the class `Car' for the KITTI dataset due to its large amount of data and complex scenarios, just as most state-of-the-art methods test their models. We follow the official KITTI evaluation protocol during the evaluation stage, and the IoU threshold is set to 0.7 for the class `Car'.

\subsection{Main Results}

\begin{table}[t]
\centering
\caption{Results on indoor datasets.}
\label{Tab11}
\begin{tabular}{c@{\ }|@{\ }c@{\ \ }c@{\ }|@{\ }c@{\ \ }c}
\toprule
\multicolumn{1}{c|@{\ }}{\multirow{2.5}{*}{Method}}                            & \multicolumn{2}{c|@{\ }}{SUN RGB-D} & \multicolumn{2}{c}{ScanNetV2} \\
                            \cmidrule(lr){2-5}
 & mAP@0.25       & mAP@0.5       & mAP@0.25      & mAP@0.5     \\ \midrule
VoteNet~\cite{qi2019deep}& 57.7           & --             & 58.6          & 33.5        \\
Reimpl.~\cite{mmdet3d2020}  & 59.1           & 35.8          & 62.9          & 39.9        \\
VoteNet+Ours               & 61.7           & 38.3          & 65.1          & 43.0        \\ \cmidrule(lr){1-5}
GroupFree3D~\cite{liu2021group}                & 63.0           & 45.2          & 69.1          & 52.8        \\
GroupFree3D+Ours            & \textbf{64.3}           &\textbf{47.5}          & \textbf{70.9}          & \textbf{54.1}     \\
\bottomrule
\end{tabular}
\end{table}

\subsubsection{Results on Indoor Datasets}
For indoor datasets, we evaluate our method on VoteNet \cite{qi2019deep} and GroupFree3D \cite{liu2021group}. VoteNet is a classic and representative 3D object detection method, while GroupFree3D is the state-of-the-art method on indoor datasets. Results are presented in Table \ref{Tab11}.
The results show that our method significantly improve both frameworks.
Compared with the baseline of VoteNet, our method achieves performance gains of 2.6\% on the SUN RGB-D with mAP@0.25 and 2.5\% with mAP@0.5. As for ScanNetV2, our model achieves performance gains of 2.2\% and 3.1\% on mAP@0.25 and mAP@0.5, respectively. Similarly, our method works for Group-Free 3D, which achieves performance gains of 1.8\% mAP@0.25 and 1.3\% mAP@0.5.

\subsubsection{Results on the Outdoor KITTI Dataset}
To fully illustrate the generalization of our approach, we have added our module to 3DSSD~\cite{yang20203dssd} and PointPillars~\cite{Lang_2019_CVPR} and carried out experiments on the KITTI dataset. The comparison results on the KITTI test set are shown in Table~\ref{kitti}. Compared with the baseline, our LG3D outperforms its original version. In terms of the main metric, i.e., AP on ``moderate'' instances, our method outperforms PointPillars and 3DSSD by 2.11\% and 1.9\%, respectively.

\begin{table}[t]
\caption{Results on the KITTI dataset.}
\label{kitti}
\centering
\begin{tabular}{c@{\ }|@{\ }c@{\ \ }c@{\ \ }c}
\toprule
\multicolumn{1}{c|@{\ }}{\multirow{2.5}{*}{Method}}  & \multicolumn{3}{c}{{AP\_3d(\%)}}          \\ \cmidrule(lr){2-4}
                                 & Easy           & Moderate       & Hard           \\ \midrule
PointPillars                     & 82.58          & 74.31          & 68.99          \\
PointPillars+LG3D                & \textbf{84.38} & \textbf{76.42} & \textbf{69.88} \\ \cmidrule(lr){1-4}
3DSSD                            & 88.36          & 79.57          & 74.55          \\
3DSSD+LG3D                       & \textbf{88.96} & \textbf{81.47} & \textbf{76.72} \\ \bottomrule
\end{tabular}
\end{table}

\subsection{Ablation Studies}
In this section, we discuss the designed choices in LG3D and investigate their independent impact on final metrics in ablation studies. If not specified, all models are designed on VoteNet of ScanNetV2.

\begin{table}[t]
\centering
\caption{Ablation study results of the LKM and LAI modules.}
\label{Tabab}
\begin{tabular}{c@{\ }|@{\ }c@{\ \ }c@{\ }|@{\ }c@{\ \ }c}
\toprule
 & \multicolumn{2}{c|@{\ }}{SUN RGB-D} & \multicolumn{2}{c}{ScanNetV2} \\ \cmidrule(lr){2-5}
                            & mAP@0.25       & mAP@0.5       & mAP@0.25      & mAP@0.5     \\ \midrule
Baseline (Reimpl.)          & 59.1           & 35.8          & 62.9          & 39.9        \\  \cmidrule(lr){1-5}
Baseline+LKM                & 61.1           & 37.3          & 64.0          & 41.3        \\
Baseline+LKM+LAI            & \textbf{61.7}  & \textbf{38.3} &\textbf{65.1}  & \textbf{43.0}   \\    \bottomrule
\end{tabular}
\end{table}

\subsubsection{Label-Guided Module} To better understand the role of our method, we conduct experiments to evaluate the contribution of each sub-task. Specifically, our LKM module comprises a label point clouds encoder and a supervision loss $L2$-distance. As shown in Table \ref{Tabab}. Even if the LKM module alone is used to supplement the original representation with label point clouds, our method achieves some performance gains. When the LAI module is used to further complement the label point clouds representation, our method achieves a further performance improvement, which shows that our main modules significantly contribute to the overall network.
For more ablation studies on label annotation augmentation strategies, please refer to the appendix.

\subsubsection{Two Steps Training}
In our method, we use a two-step training strategy. First of all, we load the well-trained function $\phi^{*}\left(\bm{P} ; \bm{\theta_{P}}\right)$, but do not freeze its parameters. We use a joint optimization method to optimize it together with $\psi(\bm{\hat{P}} ; \theta_{\mathbb{\hat{P}}})$ and $\beta\left(\mathcal{A} ; \theta_{\mathcal{A}}\right)$. By the first step, we obtain optimized $\phi^{*'}\left(\bm{P} ; \theta_{\bm{P}}\right)$, $\psi^{*}(\bm{\hat{P}} ; \theta_{\hat{\bm{P}}})$ and $\beta^{*}\left(\mathcal{A} ; \theta_{\mathcal{A}}\right)$. In the second step, we load the functions obtained in the first step, freeze all parameters, and perform the second training step to obtain the final optimized $f\left(\cdot; \theta_{f}\right)$ and $d\left(\cdot ; \theta_{d}\right)$. We show the ablation in Table \ref{Tabstage}.

\begin{table}[t]
\centering
\caption{Ablation study results of the two-stage training strategy.} \label{Tabstage}
\begin{tabular}{c@{\ }|@{\ }c@{\ \ }c}
\toprule
Method    & mAP@0.25 & mAP@0.5 \\
\midrule
Baseline  & 62.9     & 39.9   \\
One-Stage & 64.4    & 42.4  \\
Two-Stage & \textbf{65.1}    & \textbf{43.0}\\
\bottomrule
\end{tabular}
\end{table}

\subsubsection{Training with More Epochs}
The performance gain is from the proposed module, not the long training epochs. To verify that, we conduct additional experiments to train both methods with equivalent epochs (72 on VoteNet and 160 on GroupFree3D). The results in Table~\ref{tab9} indicate that training baseline detectors for a long time are not helpful, which validates the effectiveness of our approach.

\begin{table}[t]
\centering
\caption{Performance comparisons with different numbers of training epochs.}\label{tab9}
\begin{tabular}{c@{\ }|@{\ }c@{\ }|@{\ }c@{\ \ }c@{\ }|@{\ }c}
\toprule
{Dataset}          & \multicolumn{1}{c|@{\ }}{Method}              & {mAP@0.25} & {mAP@0.50} & {\# of Epochs} \\ \midrule
\multirow{7.5}{*}{ScanNetV2}  & \multirow{2}{*}{VoteNet}     & 62.90             & 39.90             & 36             \\
                          &                              & 62.50             & 40.10             & 72             \\ \cmidrule(lr){2-5}
                          & {VoteNet+LG3D}                 & \textbf{65.10}             & \textbf{43.00}             & 72             \\ \cmidrule(lr){2-5}
                          & \multirow{2}{*}{GroupFree3D} & 69.10             & 52.80             & 80             \\
                          &                              & 68.50             & 52.80             & 160            \\ \cmidrule(lr){2-5}
                          & {GroupFree3D+LG3D}             & \textbf{70.90}             & \textbf{54.10}             & 160            \\ \cmidrule(lr){1-5}
\multirow{3.5}{*}{SUN RGB-D} & \multirow{2}{*}{VoteNet}     & 59.10             & 35.80             & 80             \\
                          &                              & 59.20             & 35.70             & 160            \\ \cmidrule(lr){2-5}
                          & {VoteNet+LG3D}                 & \textbf{61.70}             & \textbf{38.30}             & 160            \\
                          \bottomrule
\end{tabular}
\end{table}

\begin{figure}[h]
  \centering
  \includegraphics[width=.95\columnwidth]{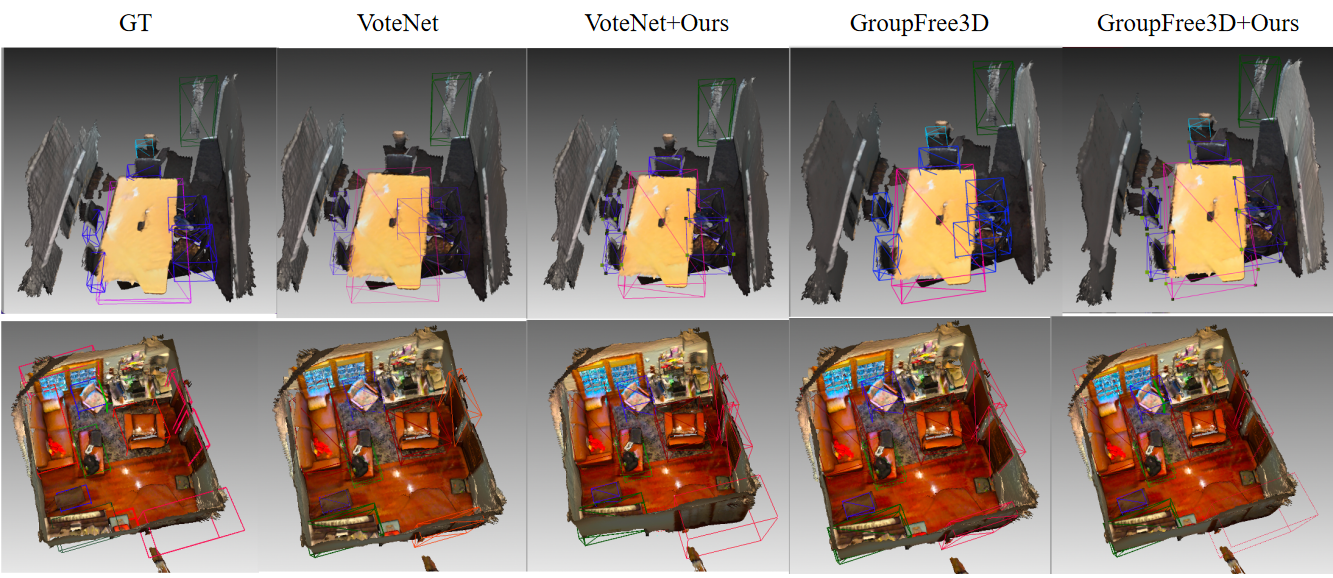}
  \caption{Result visualizations on the ScanNetV2 dataset. GT means ground truth. The bounding box color denotes the object category.}
  \label{fig:Vis}
\end{figure}

\begin{figure}[h]
\centering
\includegraphics[width=.9\textwidth]{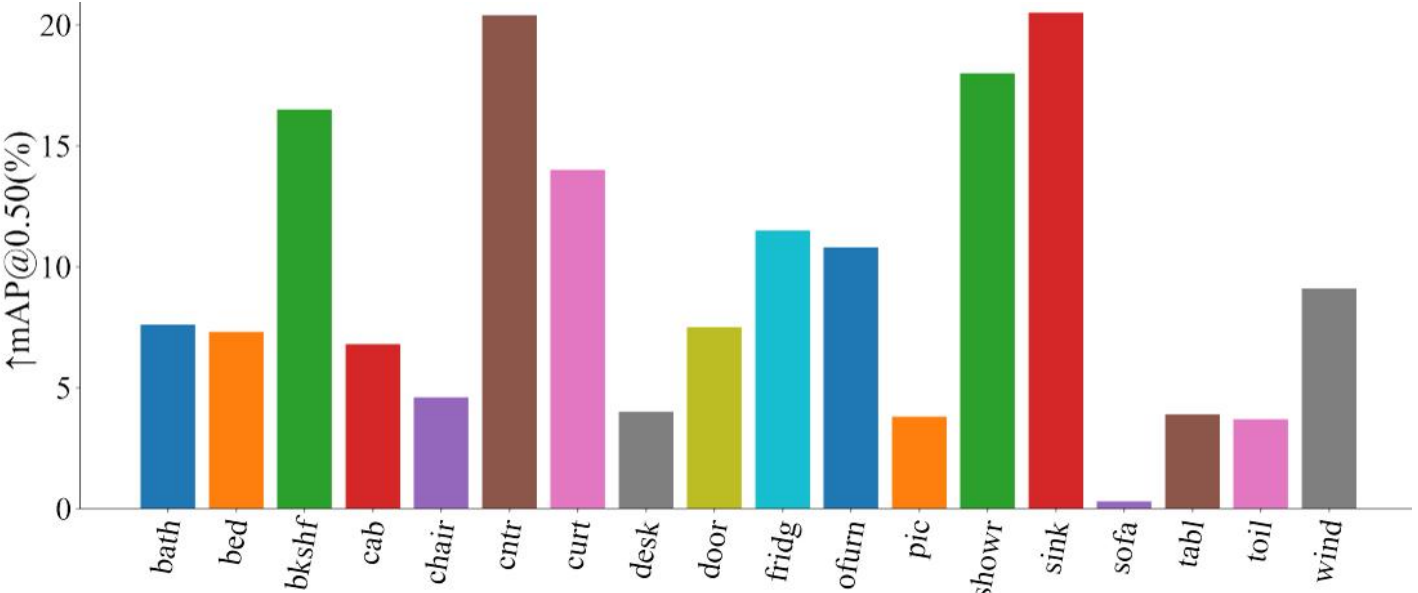}
\caption{The mAP@0.5 score improvement by LG3D applied to VoteNet of each category on the  ScanNetV2 dataset.}\label{fig:small}
\end{figure}

\subsection{Qualitative Results and Discussion}
Fig. \ref{fig:Vis} shows several representative results on ScanNetV2. Our method has a good improvement effect on the detections of missing small objects and imprecision large objects.
Fig. \ref{fig:small} shows that our method is particularly effective in detecting small objects and has an improvement for other objects.
In addition, due to the limitation of the ball query radius during feature processing, the perceptual field is naturally limited, especially for objects with large aspect ratio differences, such as shower curtains and curtains. The initial VoteNet is affected by the backbone's natural limitation and the clustering operation in the detection head part by presetting the 3D spherical boundary. Our LAI module can effectively complement the effect of this problem by using the size information in the label annotations. The results show that the performance on shower curtains and curtains in the ScanNetV2 dataset are improved by about 15\%.
For more results on per-category performance comparisons, please refer to the appendix.

\section{Conclusion}
In this work, we have designed a novel label-guided auxiliary approach for 3D object detection networks to facilitate the training process. A novel point clouds label encoding module is introduced to map real labels into potential embeddings that serve as auxiliary intermediate supervision of the detection backbone during training. A knowledge refinement-like idea is used to simplify our auxiliary module.
The processed label information is fed into the upper branch auxiliary network for encoding in the experiments.
The distance information represented by the 3D features is used to directly optimize the feature embedding of the detection backbone.
Experiments show that this method greatly improves the detection performance of the original network while maintaining the detection speed of the original network.

\section*{Acknowledgement}
This work was done when Yaomin Huang and Xinmei Liu took internships at Midea Group.
This work was supported in part by National Science Foundation of China (61731009 and 11771276) and Shanghai Pujiang Program (21PJ1420300).

\clearpage
\bibliographystyle{splncs04}
\bibliography{egbib}

\clearpage
\appendix
\counterwithin{figure}{section}
\counterwithin{table}{section}

\section{Appendix}
\subsection{Label Annotation Augmentation}
As aforementioned in Sec \ref{sec:3.2}, the proposed approach strengthens the annotations by replacing the exact size of an instance object with an uncertain size. Furthermore, the markup encoding module can identify instances and be aware of the uncertainty. We ablate their usage as shown in Table \ref{Tabdrop}. As expected, both information augments methods are helpful. Table \ref{Tabsca} shows the performance of size augment with different scale factors $B_\eta$. An appropriate scale factor allows the label encoding module to better perceive the uncertainty without losing the original annotated information.

\begin{table}[h]
\caption{Ablation study results of label information augmentation strategies.}
\centering
\label{Tabdrop}
\begin{tabular}{c@{\ \ }c@{\ }|@{\ }c@{\ \ }c}
\toprule
Size Aug. & Fake Instance & mAP@0.25 & mAP@0.5 \\
\midrule
            &               & 64.5     & 42.4    \\
\checkmark  &               & 64.9     & 42.9    \\
            &   \checkmark  & 64.8     & 42.7    \\
\checkmark  &   \checkmark  & \textbf{65.1}     & \textbf{43.0}  \\
\bottomrule
\end{tabular}
\end{table}

\begin{table}[h]
\centering
\caption{Results of label information using different scale factors.} \label{Tabsca}
{
\begin{tabular}{c@{\ }|@{\ }c@{\ \ }c}
\toprule
Scale Factor & mAP@0.25 & mAP@0.5 \\ \midrule
0 (Baseline) & 64.5     & 42.4   \\
\cmidrule(lr){1-3}
0.1      & \textbf{65.1}    & \textbf{43.0}  \\
0.2      & 64.8     & 42.7   \\
0.3      & 64.7     & 42.7  \\
\bottomrule
\end{tabular}
}
\end{table}

\subsection{Per-Category Detection Results}

\begin{figure}[h]
\centering
\includegraphics[width=.9\textwidth]{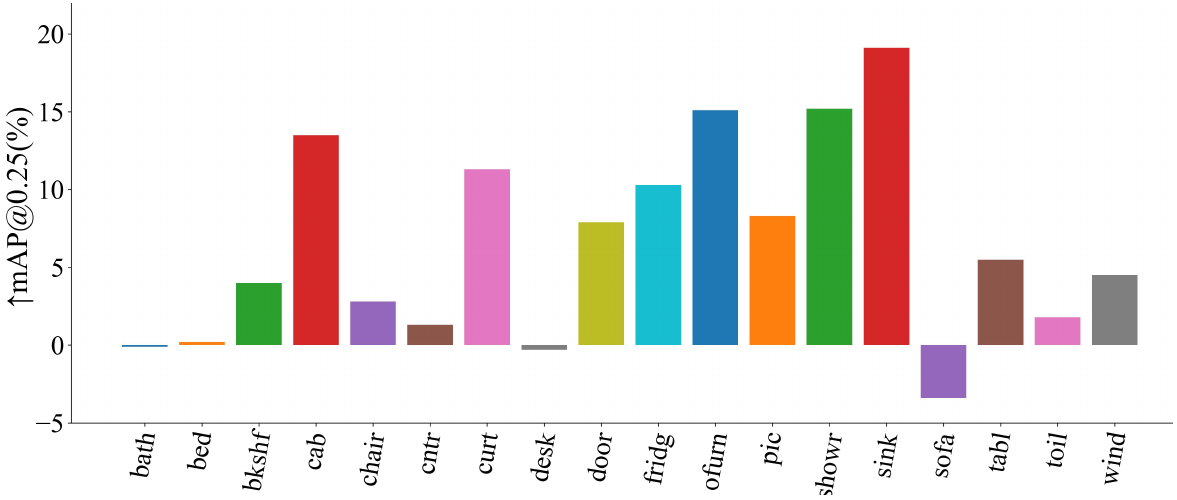}
\caption{The mAP@0.25 score improvement by LG3D applied to VoteNet of each category on the ScanNetV2 dataset.}\label{fig:small2}
\end{figure}

Our method can supplement the missing target object information in the features extracted from the original point clouds. Especially for small objects, our method has a more obvious effect. Tables~\ref{Tab4} and~\ref{Tab5} show the specific results for each category on the ScanNet.
Fig.~\ref{fig:small2} shows the improvement of LG3D on VoteNet in terms of mAP@0.25 on ScanNetV2.
From these results, we can see that the original method is less effective in detecting small objects. For example, in the picture and garbage bin categories, the original VoteNet has only 7.8\% AP@0.25 and 37.2\% AP@0.25, but after adding our module for such objects, the metrics are up to 16.1\% AP@0.25 and 52.3\% AP@0.25. This is mainly due to the complementary of our module for small object representation.

\begin{table}[h]
\centering
\caption{Per-category AP@0.25 scores on the ScanNetV2 dataset.} \label{Tab4}
\resizebox{\linewidth}{!}{
\begin{tabular}{c|cccccccccccccccccc|c}
\toprule
Method        & cab  & bed  & chair & sofa & tabl & door & wind & bkshf & pic  & cntr & desk & curt & fridg & showr & toil & sink & bath & ofurn & mAP  \\ \midrule
GSDN          & 41.6 & 82.5 & 92.1  & 87.0 & 61.1 & 42.4 & 40.7 & 51.5  & 10.2 & 64.2 & 71.1 & 54.9 & 40.0  & 70.5  & 100  & 75.5 & 93.2 & 53.1  & 62.8 \\
H3DNet        & 49.4 & 88.6 & 91.8  & 90.2 & 64.9 & 61.0 & 51.9 & 54.9  & 18.6 & 62.0 & 75.9 & 57.3 & 57.2  & 75.3  & 97.9 & 67.4 & 92.5 & 53.6  & 67.2 \\ \cmidrule(lr){1-20}
VoteNet       & 36.3 & 87.9 & 88.7  & 89.6 & 58.8 & 47.3 & 38.1 & 44.6  & 7.8  & 56.1 & 71.7 & 47.2 & 45.4  & 57.1  & 94.9 & 54.7 & 92.1 & 37.2  & 58.7 \\
VoteNet+Ours   & 49.8 & 88.1 & 91.5  & 86.2 & 64.3 & 55.2 & 42.6 & 48.6  & 16.1 & 57.4 & 71.4 & 58.5 & 55.7  & 72.3  & 96.7 & 73.8 & 92.0   & 52.3  & 65.1 \\ \cmidrule(lr){1-20}
GroupFree3D     & 52.1 & 92.9 & 93.6  & 88.0 & 70.0 & 60.7 & 53.7 & 62.4  & 16.1 & 58.5 & 80.9 & 67.9 & 47.0  & 76.3  & 99.6 & 72.0 & 95.3 & 56.4  & 69.1 \\
GroupFree3D+Ours & \textbf{56.9} & \textbf{93.1}      & \textbf{94.5}  &\textbf{89.4}     & 69.4  & \textbf{63.1}  & \textbf{55.6}& \textbf{63.6}  & \textbf{21.4} &\textbf{64.2} & \textbf{82.2} & \textbf{71.5}  &  \textbf{49.2}  & \textbf{81.4}  & 99.4 & \textbf{79.2}  & \textbf{95.8} & \textbf{61.3} &  \textbf{70.9} \\
\bottomrule
\end{tabular}
}
\end{table}

\begin{table}[h]
\centering
\caption{Per-category AP@0.5 scores on the ScanNetV2 dataset.} \label{Tab5}
\resizebox{\linewidth}{!}{
\begin{tabular}{c|cccccccccccccccccc|c}
\toprule
Method        & cab  & bed  & chair & sofa & tabl & door & wind & bkshf & pic & cntr & desk & curt & fridg & showr & toil & sink & bath & ofurn & mAP  \\ \midrule
GSDN          & 13.2 & 74.9 & 75.8  & 60.3 & 39.5 & 8.5  & 11.6 & 27.6  & 1.5 & 3.2  & 37.5 & 14.1 & 25.9  & 1.4   & 87.0 & 36.5 & 76.9 & 30.5  & 34.8 \\
H3DNet        & 20.5 & 79.7 & 80.1  & 79.6 & 56.2 & 29.0 & 21.3 & 45.5  & 4.2 & 33.5 & 50.6 & 37.3 & 41.4  & 37.0  & 89.1 & 35.1 & 90.2 & 35.4  & 48.1 \\ \cmidrule(lr){1-20}
VoteNet       & 8.1  & 76.1 & 67.2  & 68.8 & 42.4 & 15.3 & 6.4  & 28.0  & 1.3 & 9.5  & 37.5 & 11.6 & 27.8  & 10.0  & 86.5 & 16.8 & 78.9 & 11.7  & 33.5 \\
VoteNet+Ours   & 14.9 & 83.4 & 71.8  & 69.1 & 46.3 & 22.8 & 15.5 & 44.5  & 5.1 & 29.9 & 41.5 & 25.6 & 39.3  & 28.0    & 90.2 & 37.3 & 86.5 & 22.5  & 43.0 \\ \cmidrule(lr){1-20}
GroupFree3D     & 26.0 & 81.3 & 82.9  & 70.7 & 62.2 & 41.7 & 26.5 & 55.8  & 7.8 & 34.7 & 67.2 & 43.9 & 44.3  & 44.1  & 92.8 & 37.4 & 89.7 & 40.6  & 52.8 \\
GroupFree3D+Ours &  \textbf{27.7} &  \textbf{81.5} &  \textbf{83.1} & 68.2 & 61.4 & \textbf{42.6} &  \textbf{27.3} & 55.3 & 7.5  & \textbf{37.9} & \textbf{66.9} & \textbf{48.7}& 43.0    &  \textbf{46.0}  & \textbf{99.8} & \textbf{42.3} & \textbf{90.4} &  \textbf{44.7} & \textbf{ 54.1}\\
\bottomrule
\end{tabular}
}
\end{table}

\end{document}

%% file: Fig/BPC.tex
\begin{figure}[t]
  \begin{subfigure}[b]{0.32\columnwidth}
    \centering
    \includegraphics[width=\textwidth]{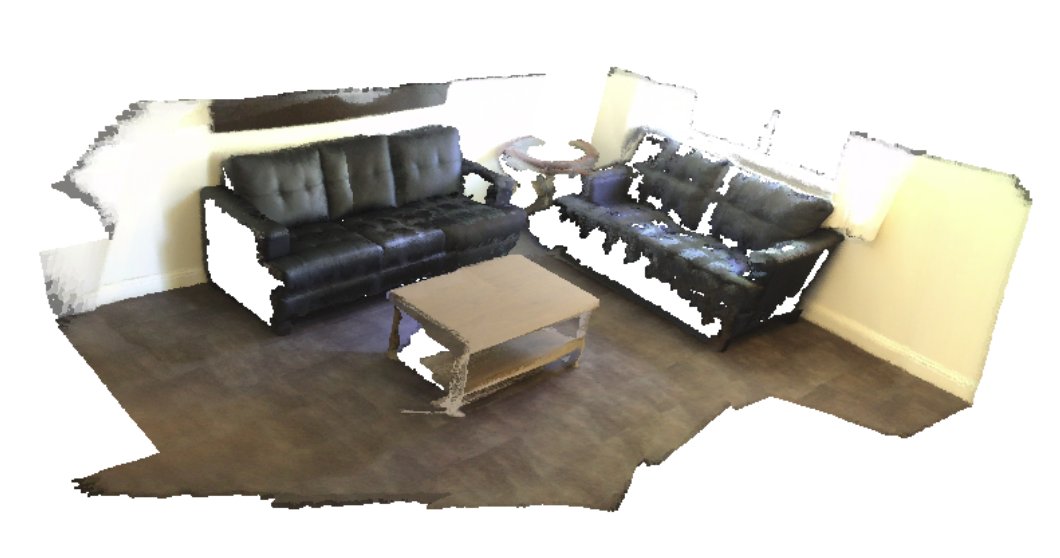}
    \caption{Point Cloud}
    \label{fig:pc}
  \end{subfigure}
  \begin{subfigure}[b]{0.32\columnwidth}
    \centering
    \includegraphics[width=\textwidth]{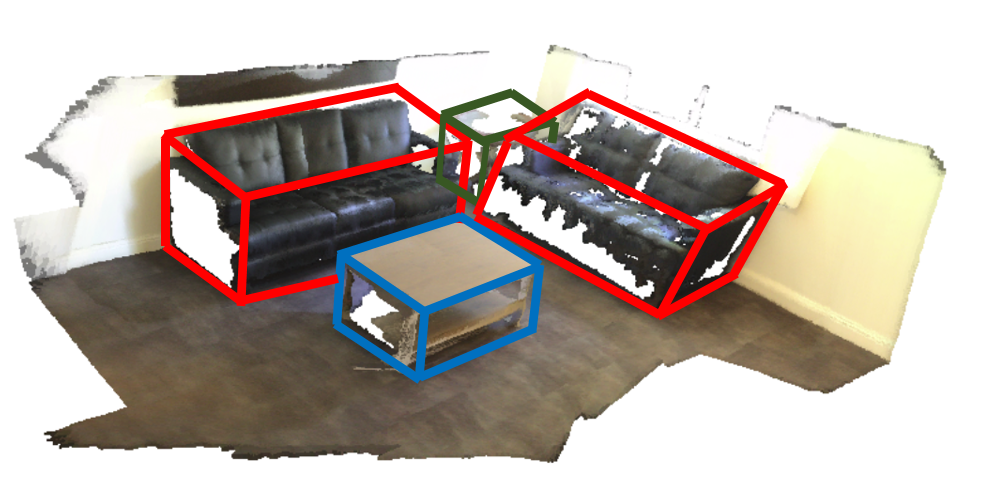}
    \caption{Bounding Boxes}
    \label{fig:bb}
  \end{subfigure}
  \begin{subfigure}[b]{0.32\columnwidth}
    \centering
    \includegraphics[width=\textwidth]{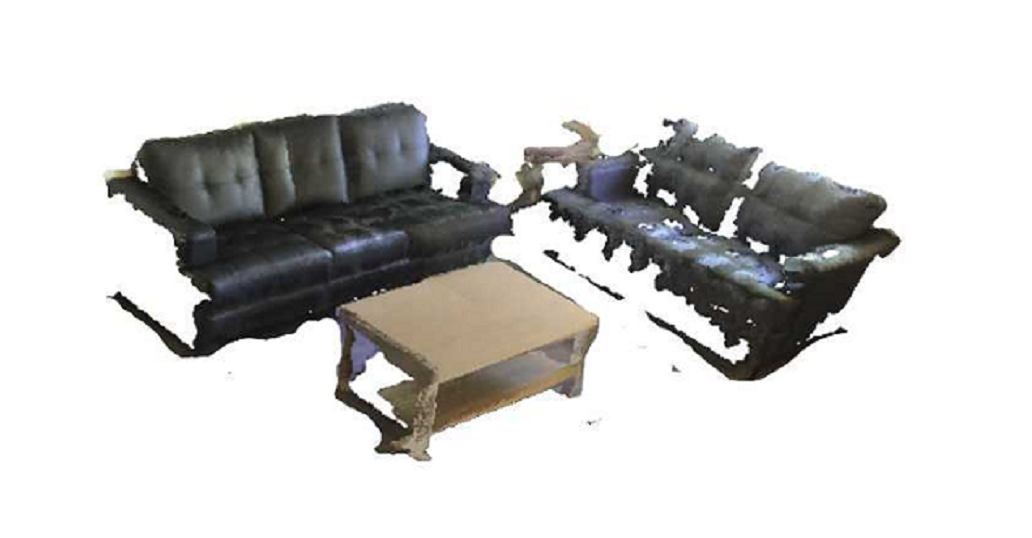}
    \caption{Labeled Point Cloud}
    \label{fig:lpc}
  \end{subfigure}
  \caption{An example of (a) original point cloud, (b) bounding box annotations, and (c) the label point cloud extracted from the annotated bounding boxes.}
  \label{fig:label}
\end{figure}